\documentclass[letterpaper, 10 pt, conference]{ieeeconf}

\IEEEoverridecommandlockouts                              

\usepackage{lipsum}
\usepackage{graphicx}                       
\usepackage{stfloats}
\usepackage[tight,footnotesize]{subfigure}  

\usepackage{amssymb,amsmath}
\usepackage{textcomp}
\usepackage{mdwmath}
\usepackage{mdwtab}
\usepackage{eqparbox}
\usepackage{rotating}                       
\usepackage{array}                          
\usepackage{listings}                       
\usepackage[ruled,vlined,noresetcount]{algorithm2e}
\usepackage{listings}
\usepackage[noend]{algpseudocode}

\usepackage[colorlinks,
            linkcolor=blue,
            anchorcolor=blue,
            citecolor=blue,
            bookmarks=false
            ]{hyperref}
\usepackage{cite}

\usepackage{graphicx}
\usepackage{float}
\usepackage{subfigure}
\usepackage{amsmath}
\usepackage{algpseudocode}
\usepackage{stfloats}
\usepackage{multirow} 
\usepackage{xcolor}
\usepackage[ruled,vlined]{algorithm2e}
\usepackage{amssymb}
\usepackage{booktabs}

\title{\LARGE \bf  An Integrated Visual System for Unmanned Aerial Vehicles Tracking and Landing on the Ground Vehicles}
\author{Kangcheng Liu$^{*}$, Xunkuai Zhou, Benyun Zhao, Huosen Ou, and Ben M. Chen
\thanks{$^{*}$Kangcheng Liu is the corresponding author.}}

\begin{document}

\maketitle
\thispagestyle{empty}
\pagestyle{empty}

\begin{abstract}
The vision of unmanned aerial vehicles is very significant for UAV-related applications such as search and rescue, landing on a moving platform, etc. In this work, we have developed an integrated system for the UAV landing on the moving platform, and the UAV object detection with tracking in the complicated environment. Firstly, we have proposed a robust LoG-based deep neural network for object detection and tracking, which has great advantages in robustness to object scale and illuminations compared with typical deep network-based approaches. Then, we have also improved based on the original Kalman filter and designed an iterative multi-model-based filter to tackle the problem of unknown dynamics in real circumstances of motion estimations. Next, we implemented the whole system and do ROS Gazebo-based testing in two complicated circumstances to verify the effectiveness of our design. Finally, we have deployed the proposed detection, tracking, and motion estimation strategies into real applications to do UAV tracking of a pillar and obstacle avoidance. It is demonstrated that our system shows great accuracy and robustness in real applications. 
\end{abstract}

\section{Introduction and Related Work}
Visual tracking and detection play a key role in all kinds of UAV navigation applications \cite{duan2015interactive}. It has a wide range of applications such as UAV search and rescue, UAV detection and tracking, UAV surveillance and environmental monitoring \cite{liu2017avoiding}, security surveillance, geographical mapping, power-line, and pipeline inspection, an autonomous inspection of large-scale bridges, warehouse management, and logistic delivery. However, the visual tracking and detection of the target objects are of great significance to improve the autonomy of unmanned aerial systems. However, some great challenges remain. First, the detection of the target object needs to be realized in real-time for the UAV. Currently, most current deep neural network-based approaches merely focus on the development of sophisticated network architecture, and pre-training algorithms to solve the detection in diverse modalities. However, the efficient algorithms which can be deployed on the UAV platform have not been sufficiently explored. Also, the robustness is poor when faced with low illuminations and rapid rotations. In order to track the UAV effectively, we need to do the motion estimation and tracking of the target object to perform the landing task.  
\begin{figure}[htbp!]
\centering
\includegraphics[scale=0.2506]{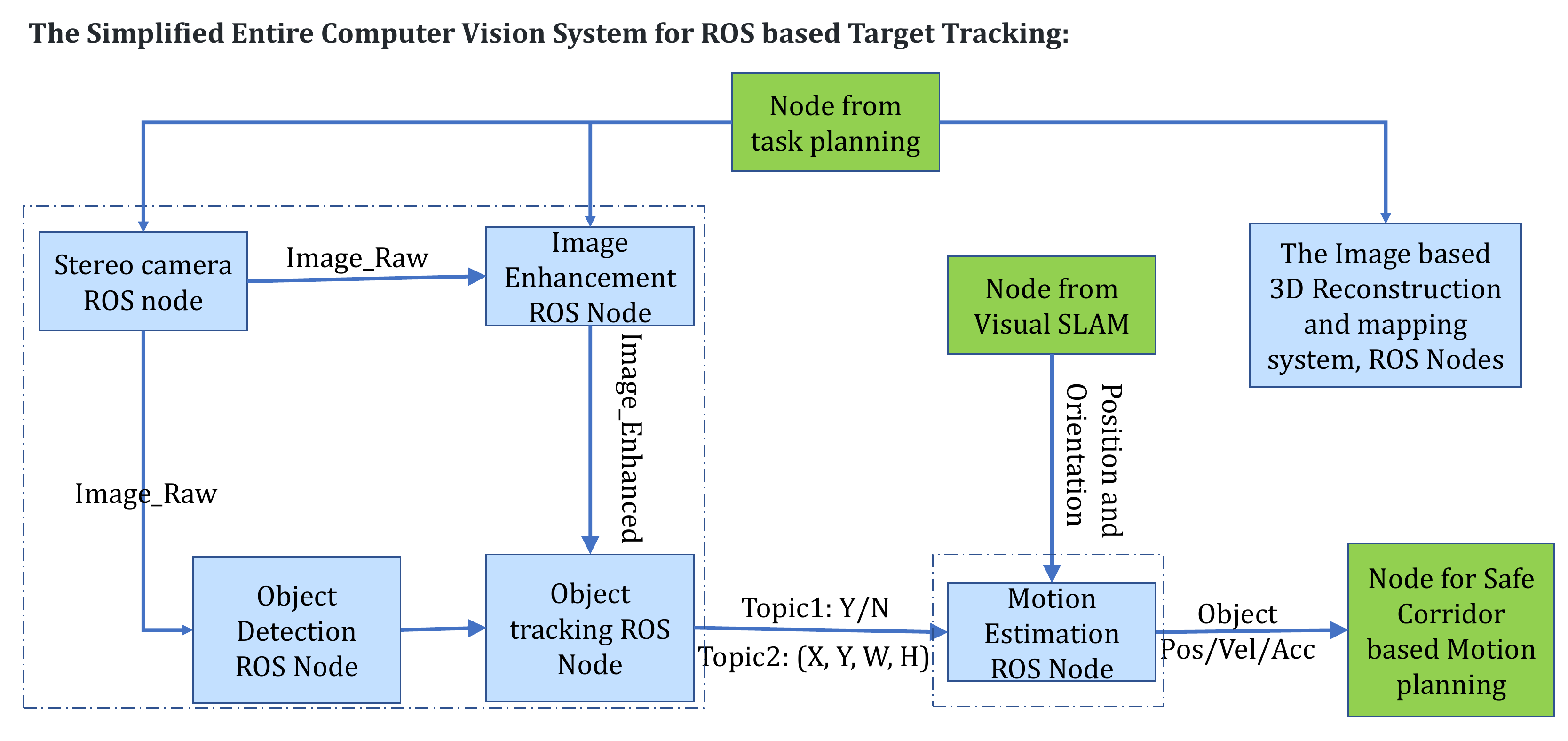}
\caption{The Detailed System Framework of the Computer Vision System for ROS-based UAV Target Tracking and Landing. The color blue indicates our proposed modules and the color green indicates other modules to fulfill the tracking and landing task.}
\label{fig_det_fram}
\vspace{-0.86cm}
\end{figure}
The Kalman Filter \cite{welch1995introduction, liu2022robustcyber, liu2022integratedicca} has been proven extensively to be an effective approach to achieving estimation of some dynamic variables given the sensor measurements observed over a period of time. However, the traditional Kalman Filter can not tackle the problem of sensor noise as well as the nonlinear motion patterns of the target. 

Yang et al. use a fuzzy logic complementary Kalman Filter (KF) based on visual and IMU data for the landing of the UAV~\cite{yang2018fuzzy}. Yuan Wei et al.~\cite{wei2020improved} use the radars installed on vehicles or the UAVs for tracking applications. Ashraf Qadir et al. use the onboard visual tracking system to implement a Kalman Filter-based visual tracking system, and the system is capable of continuously detecting the object if the tracking failure occurs \cite{qadir2011board}. But the real flight test of them remains the future work. Zhao \cite{zhao2013vision} et al. proposes a visual ground target tracking strategy for the rotorcraft UAV. Oh, et al. propose an autonomous visual tracking algorithm with Extended Kalman Filter (EKF) for micro aerial vehicles \cite{oh2007relative}. They have proposed an efficient object-tracking algorithm for UAVs, and effective ground object tracking can be achieved. Recently, various learning-based methods have been proposed for object segmentation, detection and tracking \cite{liu2020fg, liu2021fg, liu2019deep, yuzhi2020legacy, liu2022light, liu2022semi, liu2018avoiding, zhao2021legacy}, but the deep learning-based methods suffer greatly from the poor generalization capacity and the large computational and memory cost \cite{liu2022robustmm, liu2022industrialtie, liu2022weakly}. 

\begin{figure*}[htbp!]
\centering
\includegraphics[scale=0.47]{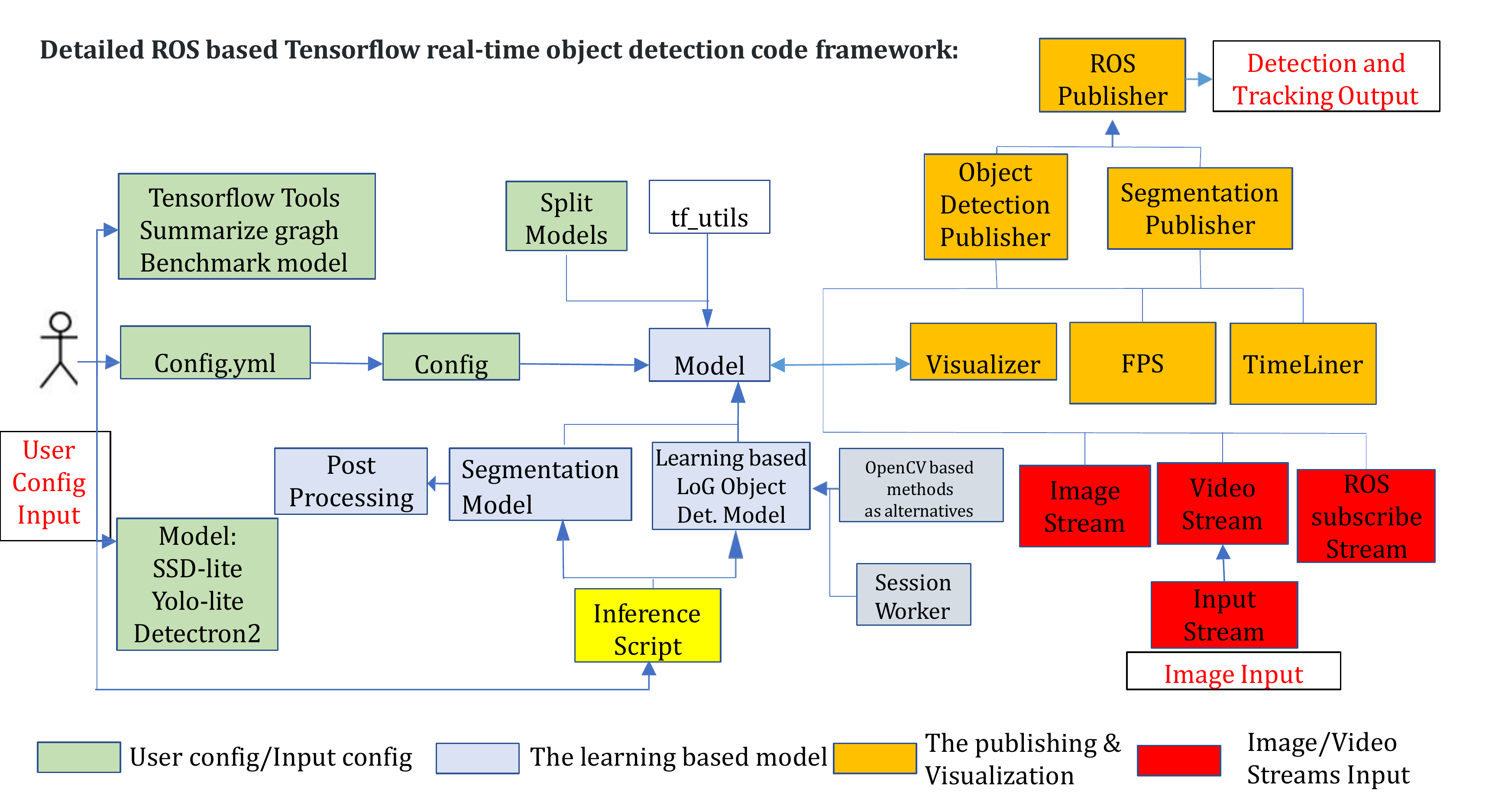}
\caption{Detailed illustration of the proposed object detection system. The green parts are the user configuration and input configuration of the detailed model. The blue part is the deep learning-based model. The orange part is for the visualization and publishing of our detection results. And the red part is for the image or video stream input.}
\label{fig_det_framework}
\vspace{-0.76cm}
\end{figure*}

In order to tackle the problems above, in the vision-based object detection and tracking, we have proposed to use a Laplacian of Gaussian filter and used it to construct a convolutional network, which makes it more appropriate for real-time object detection. It is demonstrated that our method can achieve real-time performance in an unknown environment. The recognition rate can achieve 45 frames per second, which fulfills the real-time requirements.
The UAV vision-based applications are very fundamental to all related applications \cite{liu2022light, liu2022integratedarxiv, liu2022d}. However, great challenges remain. The first is that the typical visual detection system can not handle the rotation of the targeted object and low illuminations, which makes the subsequent tracking and landing difficult. The second is that the previous Kalman filter-based motion estimation suffers from low accuracy and will greatly decrease the success rate in fulfilling the task of tracking and landing. 
As shown in Fig. \ref{fig_det_framework}, to tackle the challenges mentioned above, in this paper, we have proposed an integrated system for UAV tracking and landing applications. Taking the RGB-D images as input, we utilize our proposed Laplacian of Gaussian (LoG) filter to construct the deep neural networks to perform the object detection, which achieves robustness and accuracy under low illumination. We have proposed to use iterative multi-model methods based on the original Kalman Filter to improve the accuracy in tracking and motion estimation. Also, we have integrated our proposed approach with other robotics modules such as SLAM and motion/task planning as a whole system to perform UAV-based tracking and landing of the target objects in real applications. The deep learning-based methods have been demonstrated to be very effective in object recognition and tracking~\cite{liu2019deep, yuzhi2020legacy, liu2020fg, liu2022enhancedICCA, liu2022weaklabel3d, liu2022lightarxiv, liu2022integratedarxiv, liu2022enhancedarxiv}. In summary, we have the following prominent contributions:

\begin{enumerate}
\item We have proposed a general network for object detection and integrated it with ROS for real robotics search and rescue applications. Moreover, we have integrated the Laplacian of Gaussian (LoG) filter into the deep neural networks and it is demonstrated that the LoG-based method has a great advantage in robustness to object scale and illuminations.


\item We have done real experiments to demonstrate the effectiveness of our proposed approach. It turns out that the IMM-based filter in motion estimation shows satisfactory accuracy under various circumstances. We have also done real UAV experiments to demonstrate the effectiveness of our design.

\item We have also integrated our method with the point clouds segmentation methods for dynamic objects removal, and also we have integrated the proposed approach with motion planning approaches, which realize the real demos of the UAV landing on the UGV moving platform, as well as UAV based motion estimation of the moving/boxes pillars. 
\end{enumerate}

 \section{Proposed Methodology}
In this work, we have two prominent contributions. The first is that we proposed Laplacian of Gaussian (LoG) filter to construct the deep neural networks to perform the object detection. The second is that we have proposed to use iterative multi-model methods based on the original Kalman Filter to improve the accuracy in tracking and motion estimation. The details of these two contributions will be illustrated in detail in the remaining of this Section.

 \subsection{The Iterative Multi-model Filter for dynamic object tracking}
 \subsubsection{Kalman Filter}
 Kalman filter is an algorithm that uses linear system state equations to optimally estimate system state through system input and output observation data. Since the observation data includes the influence of noise and interference in the system, the optimal estimation can also be regarded as a filtering process. The core meaning is that the Kalman filter can estimate the state from the noisy data process very well. Moreover, the Kalman filter is also one of the breakthrough technologies used in Apollo's moon landing. Kalman filtering is also a recursive filtering algorithm based on state space in the time domain, which is easy to be realized in real time on computer, and has a small amount of computation and storage. This method can deal with the filtering problem of multi-variable non-stationary random processes and the filtering problem of time-varying systems. For example, in the course of the flight, the disturbance encountered by an aircraft is usually time-varying non-stationary noise. At this time, the Kalman filter can be used to effectively remove the interference and obtain more real state estimation data.
 To improve the detection and tracking performance, and improve the precision of positional prediction, in our application, we also develop the real-time object tracking algorithm based on the basic idea of the Kalman filter. The Kalman filter is utilized in this experiment to remove the noise in the observer and controller of the control system and minimize the number of squared errors. 
 The advantage of the Kalman filter is that it can make the optimal state estimation of the system by taking advantage of measurement data.  Kalman filter is essential because the noise and disturbance in the system influence the true data in the measurement. 
 For autonomous Unmanned aerial vehicles in our applications, denote the initial state matrix of the UAV as $X$, the initial process covariance matrix as $P$, which denotes the error in the state estimation. Next, the initial state becomes previous. Utilize subscript $K$ to represent each state in the iteration cycle. In the next time step, the current state becomes the previous one. Then the new state can be predicted based on the physical model and previous state, which can be formulated as:
 \begin{equation}
      X_{K}=AX_{K-1}+BU_{K-1}+W_{K-1}
 \end{equation}
 \begin{equation}
      Z_{k}^{'}=AP_{K-1}A^T+Q_k
 \end{equation}
The matrix $A$ is the $n \times n$ system matrix. The matrix $B$ represents the function of the input to state, which is called the input matrix or control matrix. The matrix $W_{K-1}$ denotes the predicted state noise matrix. Where the matrix $U$ denotes the control variable matrix,  $Q$ denotes the process noise covariance matrix, which keeps the state covariance matrix from becoming too small or going to zero. Denote $P_{k}^{'}$ as the prior estimation of the state, it can be represented as:
 \begin{equation}
      P_{k}^{'}=AP_{K-1}A^T+Q_k
 \end{equation}
  Let $A, B$ and $C$ denote the adaptation matrices, which convert the input state to the process state. And $Y$ denotes the measurement of state, $Z$ denotes the measurement noise. Then the measurement from sensors can be denoted as:
 \begin{equation}
      Y_{k}=CX_{K}^{\star}+Z_k
 \end{equation}  
 According to the , we can calculate the \textbf{Kalman Gain} $K$ as:
  \begin{equation}
     K=\frac{P_{k}^{'}H}{HP^{'}_{k}{H}^{T}+R}
 \end{equation} 
 Where $K$ is the \textbf{Kalman Gain}, $R$ is the sensor noise or the measurement covariance matrix. And $H$ is the conversion matrix to make the size consistent. Then the state update can be formulated as:
  \begin{equation}
     X_k= X_{k}^{'}+K(Y_k-HX_{k}^{'})
    \label{equpdate}
 \end{equation} 
 And the covariance update can be formulated as:
 \begin{equation}
     P_k= (I-KH)P_{k}^{'}
 \end{equation} 
 Where $I$ is the identity matrix. Then for the $k_{th}$ time step, the state matrix $X_k$ and the process covariance matrix which represents an error in the estimate can be obtained. Note that the adjustment of Kalman Gain is essentially the adjustment of the noise value of $Q$ and $R$. Note that:
 \begin{enumerate}
     \item The smaller the $K$ is, we can trust more on the estimation of the model.
     
     \item The bigger the $K$ is, we can trust more on the estimation of the sensors.
     
    \item The value of $K$ is related to the accuracy of the sensors and the error within the environment.
 \end{enumerate}
 It can be seen from the Eq. \ref{equpdate} that, we directly use the difference between the prediction value and the estimation value. We use the parameter $K$ to determine we trust more on the observation value $Z$ or the prediction value $X$.
 
 We apply the Kalman filter to the horizontal position and velocity of the UAV on the world coordinate. Because the x-direction and the y-direction of the horizontal coordinate are independent of each other, we merely need to set a directional state in the Kalman Filter. The state $x$ can be the position in the UAV-based visual tracking of moving object application using RGB-D camera. The position and velocity of the ground vehicle to be tracked can be set as: 
\begin{equation}       
x=\left[                 
  \begin{array}{ccc}   
    x   \\  
    vx  \\  
  \end{array}
\right]                 
\end{equation}
 The next state estimation of the vehicle can be represented as:
 
 \begin{equation}       
x^{-}_{k+1}=\left[                 
  \begin{array}{ccc}   
    x^{-}_{k+1}   \\  
    vx^{-}_{k+1}  \\  
  \end{array}
\right] = \left[                 
  \begin{array}{ccc}   
    x_{k}+vx_{k}\Delta t+w_k   \\  
    vx_{k}+w_{v,k}  \\  
  \end{array}
\right]               
\end{equation}

Then we can also obtain the system matrix $A$ from the partial derivative of $f$ with respect to $x_k$ as follows:
\begin{equation}       
A=\left[                 
  \begin{array}{ccc}   
    1 &\delta t  \\  
    0 &1 \\  
  \end{array}
\right]                 
\end{equation}
Then we can also obtain the $P_{k}^{'}$, which is the prior estimation of the state, it can be represented as:
 \begin{equation}
      P_{k}^{'}=AP_{K-1}A^T+Q_k
 \end{equation}
 Also, the  $K, X_k, P_k$ can be obtained which are the \textbf{Kalman Gain}, the state, and the covariance update. Then the motion estimation of the target UGV can be achieved.
 \begin{figure}[h]
\centering
\includegraphics[scale=0.33888]{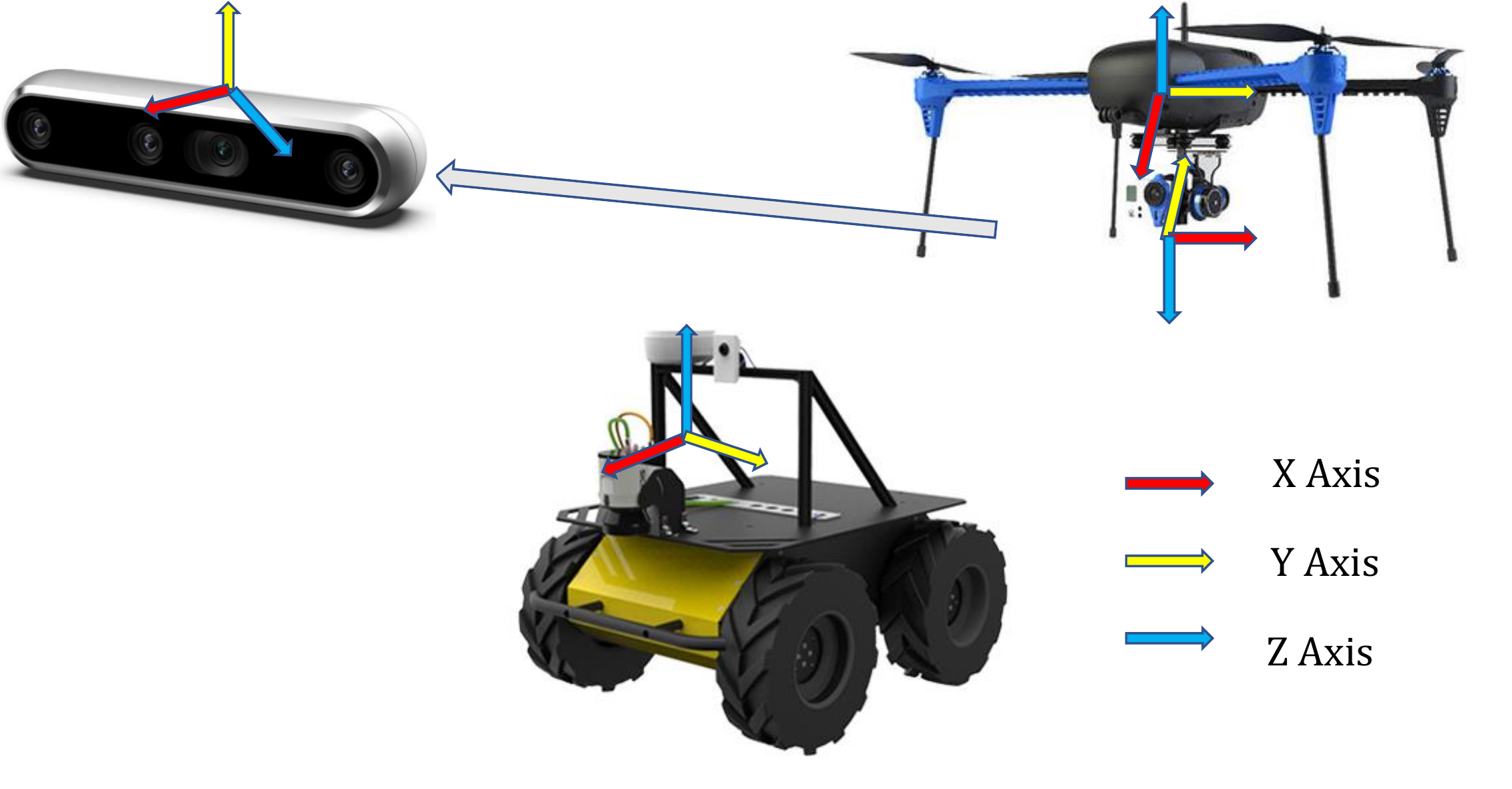}
\caption{The Coordinate Transformation Illustrations.}
\label{fig_trans}
\end{figure}
 
 \subsubsection{Proposed Iteractive Multi-model (IMM) Methods}
However, in real systems, the object may have great mobility, and it may take sudden turning and acceleration. Merely utilizing the original Kalman Filter may not realize the most ideal results, and adaptive methods must be taken. The iterative multi-model (IMM) methods \cite{li2021maximum} overcome the limitations mentioned above. The output of the filter will be the weighted average of estimation from multiple filters. The weight is the probability of correctly describing the model at the current moment. 

    


 \subsubsection{Proposed Laplacian of Gaussian-based Object Tracking Methods}
We have proposed the Laplacian of Gaussian-based object tracking results. And we have fused the Laplacian of Gaussian (LoG) operator into modern deep neural networks such as ResNext. The ResNext-based network has acceptable efficiency and we have designed methods to integrate the LoG operator seamlessly into the modern ResNext architecture. We have utilized some of our previous network designs mentioned in our FG-Net \cite{liu2020fg, liu2021fg, liu2022fg, liu2019deep, yuzhi2020legacy, yang2022datasets, liu2022ws3d}. According to our experiments, utilizing our proposed Laplacian of Gaussian operator, the performance of object detection and tracking can be significantly boosted. And we can still maintain the real-time performance and inference speed in fulfilling the tasks of aerial UAV-based object detection and tracking. 

\begin{figure}[t]
\centering
\includegraphics[scale=0.2]{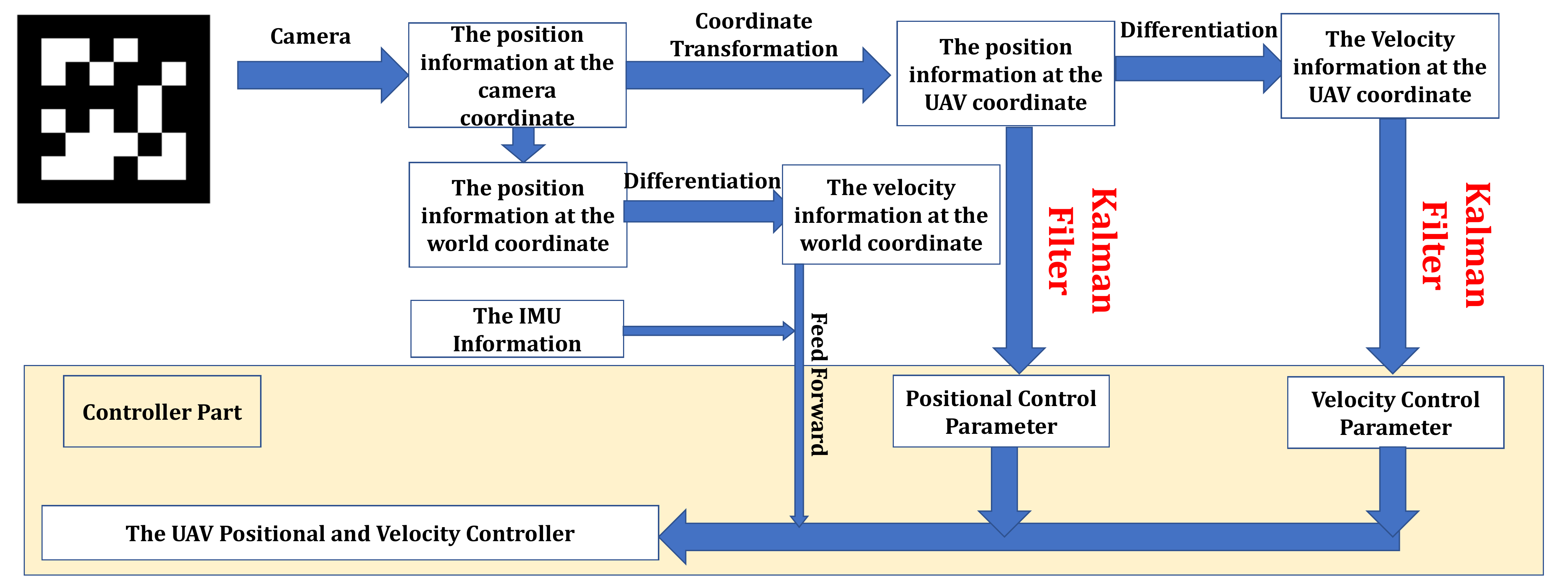}
\caption{Detailed illustration of our proposed system architecture}
\label{fig_system}
\vspace{-8mm}
\end{figure}

\begin{figure*}[h]
\centering
\includegraphics[scale=0.61888]{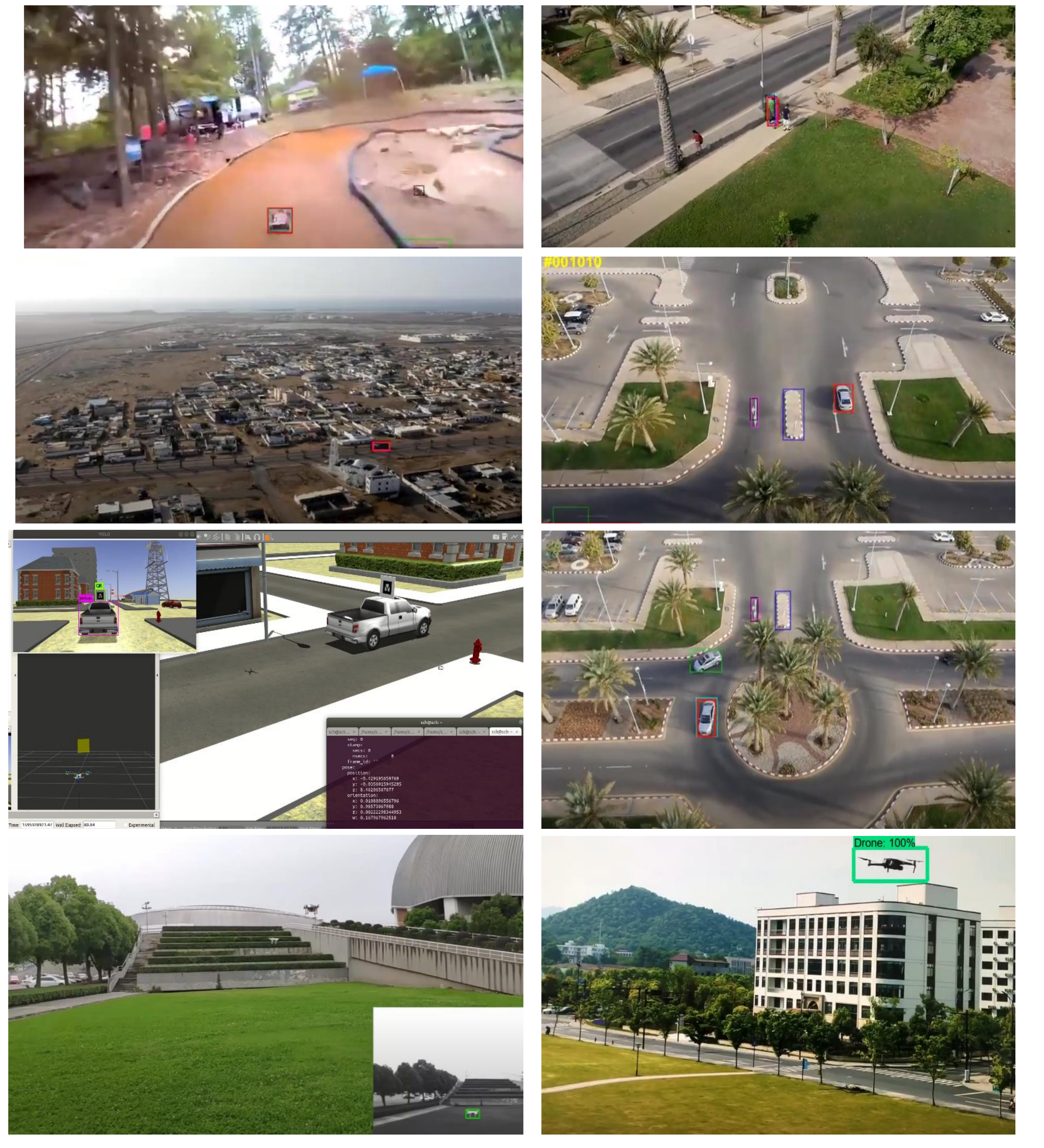}
\caption{The Real-world UAV-Based Tracking Results. The tracked objects are denoted by the bounding boxes.}
\label{fig_uav_real_track}
\end{figure*}



\begin{figure}[h]
\centering
\includegraphics[scale=0.5206]{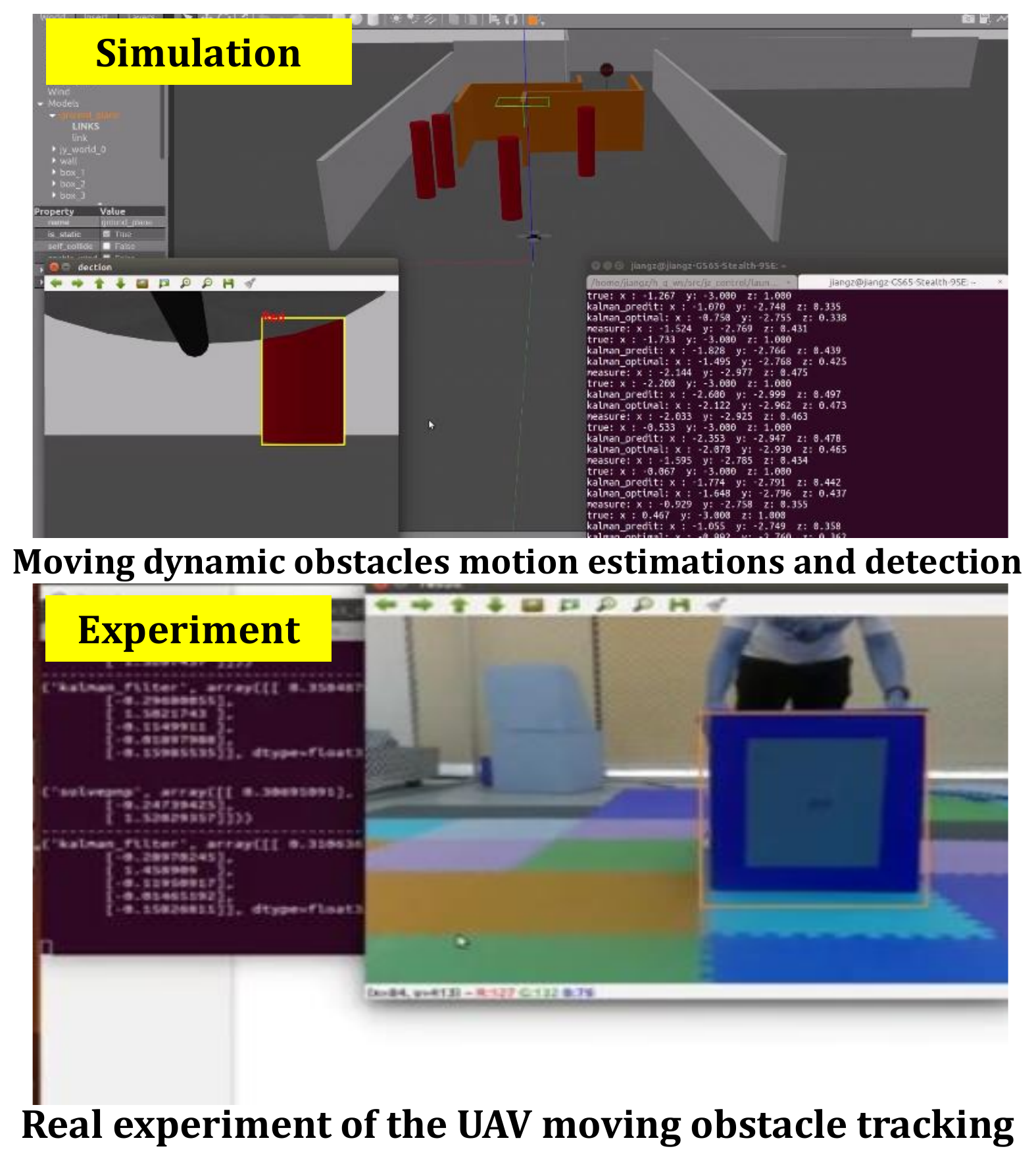}
\caption{Real-world experiments of the object tracking of the moving boxes (pillars). The figure below shows the real experiments of the UAV tracking and the figure above shows the simulation of the UAV tracking of the moving boxes (pillars). The IMM-based filter is utilized to do the motion estimation, and the LoG filter-based deep neural network is utilized to detect and track the moving pillars.}
\label{fig_uav_tracking}
\vspace{-0.2cm}
\end{figure}

\begin{table}[bp!]
\caption{The Comparisons Results of the detection performance in with fast rotation in the object box of 30 $^{\circ}$ per second.}
\label{table_rotation}
\begin{center}
\scalebox{0.96}{\begin{tabular}{ccc}
\toprule

Methods & Success & Failure in Detect\\
\hline
Ours (LoG Network)&7&0\\
Typical DCNN of ResNeXt \cite{xie2017aggregated}&5&2\\
Tradictional HOG based Method \cite{dalal2005histograms}&2&5\\

\bottomrule
\end{tabular}}
\end{center}
\vspace{-3mm}
\end{table}

\begin{table}[bp!]
\caption{The Comparisons Results of the detection performance with very low illumination.}
\label{table_illu}
\begin{center}
\scalebox{0.96}{\begin{tabular}{ccc}
\toprule

Methods & Success & Failure in Detect\\
\hline
Ours (LoG Network)&7&0\\
Typical DCNN of ResNext \cite{xie2017aggregated}&3&4\\
Tradictional HOG based Method \cite{dalal2005histograms}&1&6\\

\bottomrule
\end{tabular}}
\end{center}
\vspace{-3mm}
\end{table}
 
 \begin{table}[htbp!]
\caption{The Comparisons of The Tracking Accuracy and Computational Cost Tested in the Real scenario tracking a Moving Pillar.}
\label{table_track}
\begin{center}
\scalebox{0.726}{\begin{tabular}{ccc}
\toprule

Methods & Computational Time (ms/Image) & Max Error (cm)\\
\hline
Ours (w/o LoG based Neural Net, w/ IMM)&51.58&2.11\\
Ours (Utilized the original Kalman Filter)&75.8&3.85\\
Ours (Utilized the IMM Filter)&76.3&1.57\\
\bottomrule
\end{tabular}}
\end{center}
\vspace{-3mm}
\end{table}
 

\section{Experimental Results}
 
 

\subsection{Detection and Tracking Results}
We have summarized the related results for detection and tracking in Table \ref{table_rotation} for fast rotations and Table \ref{table_illu} for low illumination. It can be demonstrated our proposed approach has satisfactory performance under fast rotation and low illumination compared with ResNeXt-based methods and traditional methods. For our LoG based network, we adopt the same experimental settings with the state-of-the-art approach ResNext \cite{xie2017aggregated}.  It can be seen that the our proposed approach will not fail with fast rotations and very low illuminations. Therefore, the robustness of our proposed approach is demonstrated.

\subsection{Motion Estimation Performance}
We have summarized the related results of motion estimation projected onto 2D planes and the real experiments of the motion estimation of the moving pillar by the onboard Nvidia TX2 GPU on the UAV, as shown in Fig. \ref{fig_uav_tracking}. It is demonstrated by experiments that the object's position can be tracked very precisely. Also, we have compared our proposed approach and traditional Kalman Filter based approaches for motion estimation. As shown in Table \ref{table_track}, our proposed IMM filter based object tracking can realize more accurate object motion estimation. The maximum error in the motion estimation is 1.57 cm, which reduces the maximum tracking error by 2.28 cm compared with the original Kalman Filter. Also. it can be demonstrated that the LoG based neural network for detection also enhances the motion estimation accuracy. In conclusion, our proposed approach can achieve accurate motion estimation with satisfactory efficiency. As shown in Fig. \ref{fig_uav_tracking}, the real experiments of the object tracking of the moving boxes (pillars) and the related simulations are demonstrated. The figure below shows the real experiments of the UAV tracking and the figure above shows the simulation of the UAV tracking of the moving boxes (pillars). The IMM-based filter is utilized to do the motion estimation, and the LoG filter based deep neural network is utilized to detect the moving pillars.






\subsection{Final Demos of UAV landing on the Moving Platform}
Finally, we have deployed our method for a real application case of the UAV landing on the moving platform. We use the proposed network to do object detection and tracking. Based on the detection and tracking results, we use our proposed IMM filter to do the motion estimation as well as to track the target object. The simulation experiments of UAV tracking and landing on the moving UGV platform through obstacles are shown in Fig. \ref{fig_uav_real_track} and Fig. \ref{fig_uav_tracking}.  The UAV can take off and detect the objects precisely, and then track the object continuously. The UAV can autonomously approach the target, and then landing on the moving platform. Finally, the co-moving of the UAV and UGV can be achieved with great robustness.  Also, as shown in Fig. \ref{fig_uav_real_track} and Fig. \ref{fig_uav_tracking}. the UAV can land very precisely, thanks to the accurate motion estimations of position and velocity. The demos of the UAV tracking of the UGVs through various of obstacles in the road circumstances are illustrated in Fig. \ref{fig_uav_real_track} and Fig. \ref{fig_uav_tracking}. It can be demonstrated that the UAV can take off and approach the target autonomously, and then track the target UGV continuously in the obstacle. The UAV and the UGV can navigate intelligently through the obstacles. Finally, the UAV can find the way out with the UGV. The real experiments of UAV landing on the moving platform is shown in Fig. \ref{fig_uav_real_track} and \ref{fig_uav_tracking}, it can be seen that the UAV can achieve accurate and robust landing with our proposed approaches. We have compared the tracking trajectory of the UAV compared with the target ground truth trajectory. The UAV can realize very precise tracking of the target. It further demonstrates that our proposed approach can successfully fulfill the UAV tracking and landing task with satisfactory accuracy and robustness in motion estimation. 
\section{Conclusions}
In this work, we have proposed an integrated visual system for unmanned aerial vehicles landing on the targeted movingc UGV platform. We have proposed a systematical design of the vision systems for the UAV tracking and landing applications. Firstly, we have integrated the Laplacian of Gaussian filter into deep neural networks. It has shown great merits in robustness to rotation and low illumination. Secondly, we have designed an iterative multi-model Kalman filter adapted based on the original Kalman Filter for object tracking, which achieves great accuracy. Finally, we have integrated our proposed approach with other robotics modules such as SLAM and motion/task planning as a whole system to perform UAV tracking and landing in real applications. UAV Vision is important \cite{liu2022datasets, liu2022rm3d}. Our integrated visual system is very important for future UAV-based detection and tracking applications such as autonomous landing on moving vehicles.

\addtolength{\textheight}{0cm}   





\bibliographystyle{IEEEtran}
\bibliography{references}

\end{document}